\documentclass[letterpaper, 10 pt, conference]{ieeeconf}
\IEEEoverridecommandlockouts
\usepackage{amsmath} % assumes amsmath package installed
\usepackage{amssymb}  % assumes amsmath package installed
\usepackage{glossaries}
\usepackage{siunitx}
\usepackage{cancel}
\usepackage{placeins}
\usepackage[dvips]{graphicx}
\usepackage{caption}
\usepackage{subcaption}
\usepackage{colortbl}
\usepackage{xcolor}
\usepackage{url}
\usepackage{adjustbox}
\usepackage[showonlyrefs]{mathtools}
\usepackage{pifont}
\usepackage{comment}
\usepackage{booktabs}
\usepackage{mdframed}
\usepackage{adjustbox}
\usepackage{authblk}
\usepackage{color}
\usepackage{bm}
\usepackage{algorithm}
\usepackage{xurl}
\usepackage{cite}
\makeatletter
\let\NAT@parse\undefined
\makeatother
\usepackage{float}
\usepackage{pifont}
\usepackage{hyperref}
\hypersetup{colorlinks,allcolors=black}
\usepackage{algpseudocode}
\usepackage{array}

\author{Sanket A. Salunkhe$^1$$^*$, Pranav Nedunghat$^1$$^*$, Luca Morando$^1$, Nishanth Bobbili$^1$,\\ Guanrui Li$^2$, and Giuseppe Loianno$^1$
\thanks{$^*$Equal contribution.}
\thanks{$^1$The authors are with the New York University, Tandon School of Engineering, Brooklyn, NY 11201, USA. {\tt\footnotesize email: \{sas9908, pn2187, luca.morando, nb3553, loiannog\}@nyu.edu}.}
\thanks{
$^2$The author is with the Worcester Polytechnic Institute, Robotics Engineering, Worcester, MA 01609, USA. {\tt\footnotesize email: gli7@wpi.edu}.}
\thanks{This work was supported by the NSF CPS Grant CNS-2121391, the NSF CAREER Award 2145277, the DARPA YFA Grant D22AP00156-00, Qualcomm Research, Nokia, and NYU Wireless.}
}

\title{\LARGE \bf 
Intuitive Human-Drone Collaborative Navigation in Unknown Environments through Mixed Reality
}
\begin{document}

\makeatletter

\g@addto@macro\@maketitle{
\setcounter{figure}{0}
\centering
 \includegraphics[width=\textwidth]{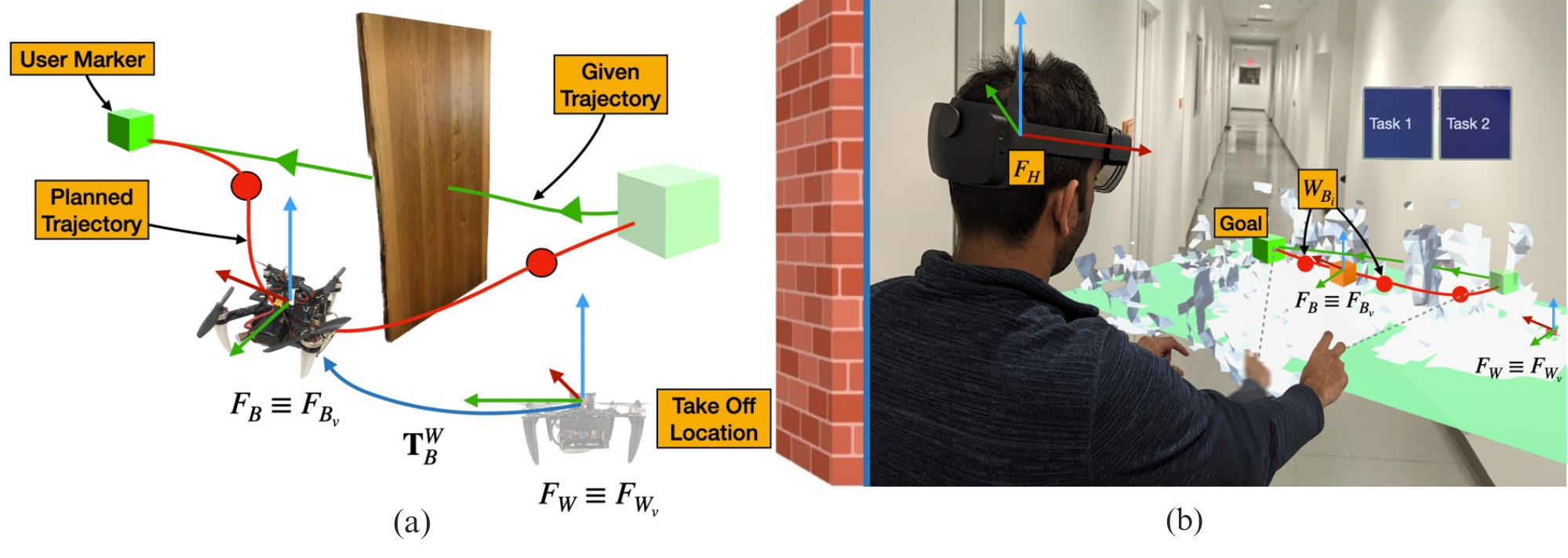}
 \captionof{figure}{Overview of the system. On the right side (b), the users assign a trajectory to the robot in the minimap using either one of the two blue task buttons present in front of them. The robot will then autonomously navigate to follow the same trajectory in the real world while avoiding obstacles, as showcased on the left side (a). With the proposed system, the user can safely operate the drone without needing to maintain direct visual contact.\label{fig:intro}}
  \vspace{-10pt}
}

\makeatother
\maketitle

\begin{abstract}
Considering the widespread integration of aerial robots in inspection, search and rescue, and monitoring tasks, there is a growing demand to design intuitive human-drone interfaces. These aim to streamline and enhance the user interaction and collaboration process during drone navigation, ultimately expediting mission success and accommodating users' inputs. In this paper, we present a novel human-drone mixed reality interface that aims to (a) increase human-drone spatial awareness by sharing relevant spatial information and representations between the human equipped with a Head Mounted Display (HMD) and the robot and (b) enable safer and intuitive human-drone interactive and collaborative navigation in unknown environments beyond the simple command and control or teleoperation paradigm. We validate our framework through extensive user studies and experiments in a simulated post-disaster scenarios, comparing its performance against a traditional First-Person View (FPV) control systems.
% Experimental results carried out in a sample post-disaster scenario, validate the proposed solution. 
Furthermore, multiple tests on several users underscore the advantages of the proposed solution, which offers intuitive and natural interaction with the system. This demonstrates the solution's ability to assist humans during a drone navigation mission, ensuring its safe and effective execution.
\end{abstract}

\vspace{-10pt}
\section*{Supplementary Material}
\noindent\textbf{Code}:
\url{https://github.com/arplaboratory/hri}

\IEEEpeerreviewmaketitle
%!TEX root = ARTICLE.tex
\section{Introduction}
Aerial robots have the potential to help humans in complex or dangerous tasks such as exploration~\cite{Burschka2012}, inspection~\cite{Ozaslan2016}, mapping~\cite{LoiannoICRA2015}, transportation~\cite{guanrui2021ral} and search and rescue~\cite{Michael_JFR}.
Enhancing human-robot interaction capabilities during challenging tasks is crucial. This enables the user to have greater control and improves their spatial awareness by complementing or augmenting their perceptual capabilities with those obtained from the robots through various sensing modalities.
%understanding of the robot's observation of its surroundings with a clear representation of the collected sensor data.
The enhanced interaction should be able to minimize potential challenges encountered by the human users during the remote robot operation, generally identified as ``context switching" \cite{DOISY2017527} or ``motion sickness" \cite{seo2017monocle}.
%One of the primary challenges involves determining the medium, type, and quantity of information that should be exchanged between humans and robots, in order to create an efficient collaboration during a shared task. 
 
Most human-robot interfaces provide the operators a basic way to send high-level commands to the robots, often relying on devices like remote controllers~\cite{Chen_TeleRobots}, joysticks~\cite{Isop2019}, and keyboards while the users also receive visual feedback from the robot through monitors~\cite{Isop2019}, LEDs, or audio signals~\cite{Magassouba}. However, these interfaces require well-trained piloting skills and limit the robot's mobility to stay within the human visual line of sight. As a result, this can restrict the range of movements of the system and the variety of tasks it can perform. This is particularly evident in real-world constrained environment missions such as search and rescue scenarios. For instance, expert firefighters, hindered by limited interaction capabilities with drones, only use these robots to hover over areas of interest, fearing that closer inspection might lead to hazardous situations. 

This work presents a deeply integrated mixed-reality human-drone collaborative interaction framework to overcome the limitations of conventional systems. This framework simplifies user interaction and communication with a companion aerial robot during navigation in unstructured and cluttered indoor environments. This promotes safer use of the robot during autonomous exploration tasks of unknown environments while concurrently decreasing the user's operational cognitive workload. 
Unlike previous approaches that merely combine independent MR and drone navigation components, our system achieves seamless bi-directional integration where the MR interface and autonomous drone navigation are inherently interdependent enhancing the user awareness by overlying the real world with digital spatial elements. This provides the user with hands-free, gesture-based control of the robot, enhancing situational awareness and spatial understanding, as visible in Fig.~\ref{fig:intro}. 
We believe that the proposed system offers a compelling opportunity to enable multi-modal information sharing within the human-robot team, potentially outclassing standard interfaces generally only based on vision \cite{HALME2018111}, gestures \cite{LIU2018355}, natural languages \cite{Pate2021}, and gaze \cite{LoiannoIcra, Pavliv2021}.

In this paper, we provide the following key contributions

\begin{itemize}
\item We present a novel bi-directional spatial representation system that enables continuous sharing of spatial data between the robot and the user, removing the need for the robot to stay within the operator’s line of sight.

% We introduce a novel bi-directional spatial representation system that allows a continuous sharing of spatial data between the robot and user, eliminating the need for the robot to remain within the operator's line of sight.

\item We tightly couple within the system the drone's autonomous navigation and re-planning capabilities with an intuitive mixed-reality interface. This setup allows users to interact with the environment and remotely set goals or trajectories for the robot. 

\item We demonstrate the effectiveness of our approach through comprehensive experiments, highlighting its advantages in a reduced mental cognitive loading over a traditional First-Person View (FPV) control method especially in unknown indoor environments. The system’s performance is cross-validated using the NASA Task Load Index (TLX) \cite{hart1988development} to assess cognitive workload alongside quantitative metrics such as the total area explored within a fixed timeframe, demonstrating significant improvements in both user experience and operational efficiency.
\end{itemize}

The remainder of this paper is organized as follows. Section \ref{sec:related-works} reviews the related works relative to our contribution. In Section \ref{sec:system_design}, we present the system design, the Mixed Reality interface, and our built-in-house UAV platform. Section \ref{sec:experimental_results} introduces the experiment results, whereas Section \ref{sec:conclusion} concludes the paper and outlines future work.

%!TEX root = ARTICLE.tex
\section{Related Works}\label{sec:related-works}

% Eliminate Hololens 2 from Figure 2
\begin{figure*}[!t]
  \centering
  \includegraphics[width=0.95\textwidth]{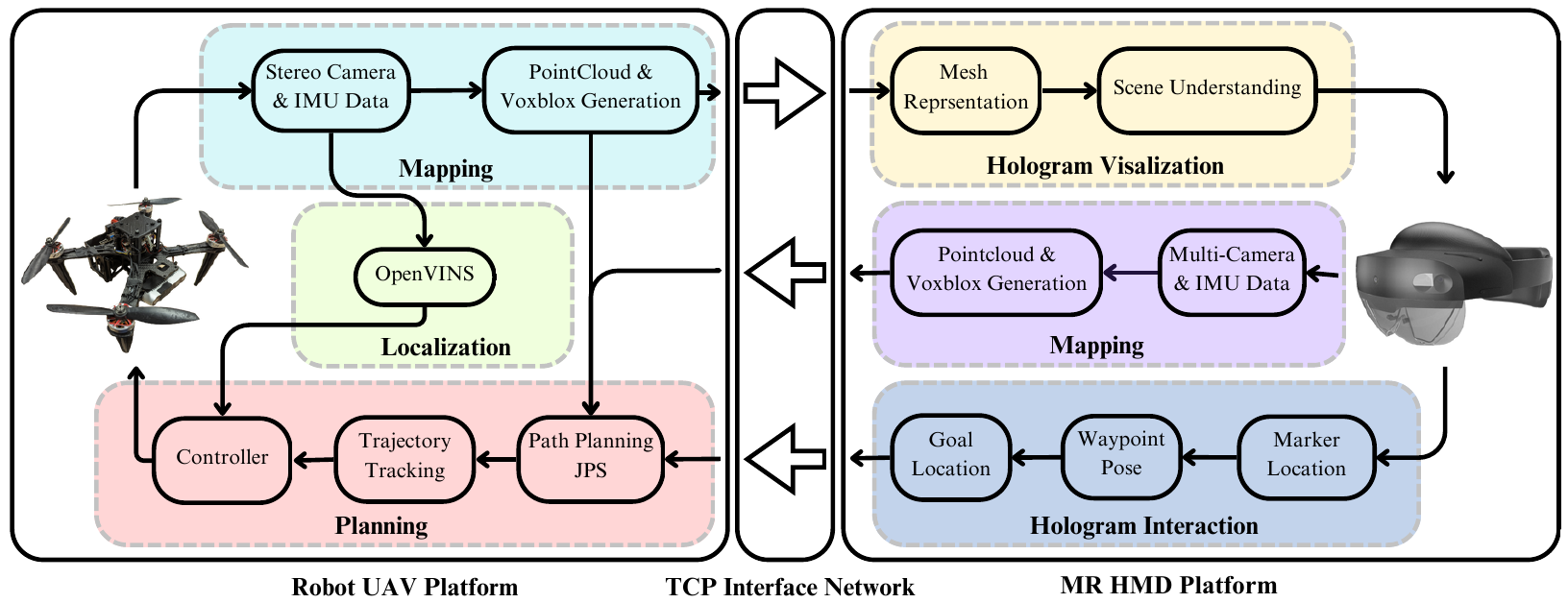}
  \caption{Block scheme describing the communication pipeline and the software architecture between the robot and the headset.   \label{fig:block_diagram}}
  \vspace{-14pt}
\end{figure*}

\textbf{Control Interfaces}. 
Researchers have started using human-drone interfaces typically based on radio controllers for velocity control or GPS setpoint commands. In most cases, during robot operation, the user relies mostly on visual feedback from the robot camera displayed on monitors. 
For example, in~\cite{Rojas2019, Isop2019}  the authors propose a ground station with a virtual representation of the environment. Although these representations are precise, the type of human-drone interaction is still largely limited to joystick-based control. 

Recent advancements try to overcome the limitations regarding the intuitiveness of teleoperation, which is a crucial factor in time-sensitive tasks, such as search and rescue missions. These innovations include impedance control with vibrotactile feedback~\cite{tsykunov2019swarmtouch} and the integration of standard joysticks with haptic force feedback. For example, in~\cite{Hala2011}, the authors propose a leader-follower-based controller enhanced with haptic force feedback derived from obstacles detected by the robot during exploration.  Different from these solutions, our interaction modality is based on extended reality techniques and does not limit anymore the drone operational range to be in close proximity to the user. 

\textbf{Extended Reality}. The recent emergence of extended reality interfaces (e.g., Augmented, Virtual, and Mixed Reality) introduces innovative, intuitive, and valuable interaction modalities, greatly improving information exchange and visualization in robotics tasks~\cite{walker2022virtual, Liu2018}. This, in turn, enables efficient communication of the robot's intentions to human teammates ~\cite{coovert2014spatial, rosen2017communicating}.
In~\cite{Erat2018,Betancourt2022}, a virtual representation of the drone's exocentric view is provided to the user who in turn can command the desired final drone's poses through intuitive pick-and-place gestures, whereas in~\cite{angelopoulos2022drone} a sequence of multiple waypoints can be assigned. 
With a similar paradigm, in~\cite{Szafir2} the authors propose an interesting Mixed Reality (MR) solution with the aim of enhancing situational awareness during human-robot collaboration, by exchanging the ground station with a miniature holographic satellite map of the target area where real status updates from the robot are represented.
Finally in~\cite{morando2024spatial}, a spatially assisted interaction method in MR is proposed, allowing users to seamlessly interact with robots using Variable Admittance Control~\cite{Keemink_admittance}, facilitating smooth joint exploration within cluttered environments.

Compared to the aforementioned solutions especially~\cite{Szafir2,morando2024spatial}, our approach uniquely combines MR with an autonomous aerial robot pipeline which allows for a continuous re-planning of the path through cluttered indoor environments, leveraging customized autonomous drone navigation and mapping capabilities, while facilitating relative information sharing with the user. 
\section{Methodology}\label{sec:system_design}

In this section, we first introduce some preliminary concepts, such as frame definition and notation. We then show how both the Head Mounted Display (HMD) and the drone perform localization and mapping in unstructured environments. Subsequently, we detail the human-drone interaction through an MR interface, enabling the user to guide the drone toward a specified spatial goals or follow a given trajectory. The overview of system design is shown in Fig.~\ref{fig:block_diagram}.

\subsection{Preliminaries}
In our setup, as shown in Fig.~\ref{fig:intro}, we employ the following coordinate systems
\begin{itemize}
    \item $\mathcal{F}_{W}$: Inertial frame of robot.
    \item $\mathcal{F}_{B}$: Robot body frame.
    \item $\mathcal{F}_{H}$: HMD body frame.
    \item $\mathcal{F}_{B_v}$: Frame attached to the virtual robot.
    \item $\mathcal{F}_{M_v}$: Frame attached to the interactive marker.
    \item $\mathcal{F}_{W_v}$: Common frame between Virtual World and Robot Inertial frame. 
\end{itemize}
We denote $\mathbf{T}_i^j$ as a generic transformation matrix from frame $i$ to frame $j$. We define the following states for the drone  
$\begin{bmatrix}
{\mathbf{x}^W}^\top & \dot{\mathbf{x}}^{{W}^\top} & {\mathbf{q}_B^W}^\top & {\bm{\omega}_B}^\top \end{bmatrix}^\top$, 
where $\mathbf{x}^W$ and $\dot{\mathbf{x}}^W$ denote, respectively, the position of the robot and the linear velocity expressed in the inertial frame $\mathcal{F}_{W}$, $\mathbf{q}_B^W$ denotes the robot attitude in quaternion with respect to $\mathcal{F}_{W}$ and $\bm{\omega}_B$ is the robot's angular velocity with respect to its body frame $\mathcal{F}_{B}$.

\subsection{Localization, Mapping, and Data Fusion}
\label{sub:Loc and Mapping}
\subsubsection{Drone}
The robot leverages onboard sensors and algorithms for state estimation, which allows the drone to autonomously navigate in cluttered environments. 
It is equipped with onboard sensors as depicted in Fig.~\ref{fig:block_diagram}. The front stereo camera~\cite{Thakur_Loianno_Liu_Kumar_2020, guanrui2020planning} generates a point cloud of the surrounding environment for mapping purposes. The point cloud $P_{B}^{W}$ is initially defined in the robot's body frame $\mathcal{F}_{B}$ and then transformed into the inertial frame $\mathcal{F}_{W}$ using transformation $\mathbf{T}_{B}^{W}$, obtained using the robot state estimation. This point cloud data is used to create a geometric representation of the surrounding environment through a real-time Octomap representation, denoted as $V_{B}^{W}$, useful for robot autonomous planning and for enhancing user situational awareness. Leveraging the Neural Euclidean Signed Distance Field~\cite{ortiz2022isdf} spatial information provided by the mapping algorithms, the low-level path planning algorithm described in Section~\ref{sec:assisted_nav} can navigate to any desired location in the scenario, generating a collision-free path to the goal.

\subsubsection{HMD}
Consumer-grade HMD MR devices offer a promising platform to create novel human-robot interaction paradigms. They provide a lightweight, hands-free, and mobile solution that supports continuous spatial information sharing and enables gesture-based interaction.

In order to share the data between the robot and the HMD, we introduce a common frame denoted as $\mathcal{F}_{W_V}$, which is represented as a holographic object in the virtual world and accurately positioned by the user to align with the robot $\mathcal{F}_{W}$  before the take-off as shown in Fig.~\ref{fig:uav_path_result}.
At this point, the robot's point cloud ${P}^W_{robot}$ can be accurately integrated and aligned with the real-world as seen through the HMD. Additionally, the drone's estimated pose $\mathbf{x}^W,\mathbf{q}^W$ is consistently shared with the HMD and localized within the generated map. This gives the user real-time visual position feedback of the robot, even when not directly within the line of sight.
To better understand the operational space from the users' perspective, we transform the volumetric voxel representation into a smooth mesh. This mesh, represented as $M_{robot}^{W_V}$, illustrates the robot's map surfaces, and it is presented in MR as a hologram overlayed to the real-world representation as shown in Fig.~\ref{fig:uav_path_result}.a or as a scaled-down version of the hologram as seen in Fig.~\ref{fig:intro}.b. The resulting virtual map can be displayed with real or scaled dimensions, focusing the user's attention only on relevant data and displaying a real-time feedback update of the robot's exploration progress.  
Lastly, we leverage the headset generated point cloud,  $P_{holo}^{H}$ expressed in $\mathcal{F}_{H}$, to ensure bidirectional and consistent spatial information sharing between the user and robot. The user-generated point cloud $P_{holo}^{H}$ is transformed into the robot's inertial frame $\mathcal{F}_{W}$ where it is merged with the robot's point cloud $P_{robot}^{W}$ to generate a joint and spatially consistent Octomap geometric representation.

\begin{algorithm}[!t]
\caption{Robot Path Planning through MR. \label{alg:path_planning}} 
\renewcommand{\algorithmicrequire}{\textbf{Input:}}
\renewcommand{\algorithmicprocedure}{\textbf{Thread:}}
\renewcommand{\algorithmicensure}{\textbf{Output:}}
\begin{algorithmic}
% \Procedure{A. Localization}{}

\While{User drawing trajectory  $\bm{\sigma}_{d_{W_v}}$ in minimap}
    \State Sample $\bm{\sigma}_{d_{W_v}}$ into $\mathcal{W}_{B,1}^{W_v}, \cdots, \mathcal{W}_{B,n}^{W}$ every $t~\si{Sec}$
    \For{$i = 1\cdots n$} 
    \State Scale $\bm{\sigma}_{d_{W_v, i}}$ using minimap scale factor 
    \State Transform $\bm{\sigma}_{d_{W_v, i}}$ to $\bm{\sigma}_{d_{W,i}}$ using $\mathbf{T}_{W_v}^{W}$
    \State Append $\bm{\sigma}_{d_{W,i}}$ to  $\bm{\sigma}_{d_{W}}$
    \EndFor
\EndWhile

\If{User command "Publish"}
    \State Send $\bm{\sigma}_{d_{W}}$ to JPS Planner
    \State Transform $\bm{\sigma}_{d_{W}}$ into $\bm{\sigma}_{d_{B}}$ using $\mathbf{T}_{W}^{B}$
    \For{$\bm{\sigma}_{d_{B,i}}$ in $\bm{\sigma}_{d_{B}}$}
        \State Construct occupancy grid from $P_{robot}^{W}$
        % \State Map $\overrightarrow{pose_B}(x,y,z)$ to occupancy grid $voxel(target)$
        \If{$\bm{\sigma}_{d_{B,i}}$ is in $obstacle$}
            \State $\bm{\sigma}_{d_{B,i}} = Nearest Free Pose$
        \EndIf
        \State $\bm{\sigma}_{r_{B,i}}$ = JPS($\bm{\sigma}_{d_{B,i}}$)
        \State $TrajectoryTracking(\mathcal{W}_{B}^{W})$
    \EndFor
\EndIf 

\end{algorithmic}
\end{algorithm}
\vspace{-10pt}

\subsection{Navigation Modalities}
\subsubsection{MR-based Autonomous Drone Navigation}
\label{sec:assisted_nav}
As previously discussed, the HMD-MR provides versatile drone interaction modes, presenting augmented surroundings either in full scale, shown in Fig.~\ref{fig:uav_path_result}.a, or as a miniaturized ``Minimap", shown in Fig.~\ref{fig:intro}.b. 
Regardless of the selected scale, users can set new autonomous goals using hand gestures and spatial buttons named "Task 1" and "Task 2". Specifically, upon pressing "Task 1", the green cube above the virtual map becomes interactive, allowing users to position it as the drone's destination. 
Pressing ``Task 2" presents a similar graphic interface but with a distinct drone behavior. 
The green cube not only designates the final goal but, when dragged along the map, also outlines a desired sample trajectory $\bf{\sigma}_{d_{W}}$ in $\mathcal{F}_W$. 
%In both tasks, the desired commands are sent to the drone only after the ``Publish" voice command is given, triggering the autonomous pipeline and activating the drone from a hovering state.

In both interactions, the users do not need to design an obstacle-free trajectory for the robot, as the robot's planner incorporates a re-plan mechanism. In this way, this feature can override the user's desired command ($\bf{\sigma}_{{d}_W}$) to obtain a safer, collision-free re-planned path ($\bf{\sigma}_{{r}_W}$), ensuring the robot reaches its assigned goal while also compensating for potentially incorrect or unsafe user motions.
To achieve this, the autonomy pipeline employs the Jump Point Search (JPS)~\cite{Harabor_Grastien_2014} algorithm for collision-free path generation. 
%In cases where the user-defined waypoints fall within an occupied or unsafe region, $NearestFreePose$ function identifies the closest position in free space. By doing so, it ensures that even user-specified trajectories are continuously adjusted dynamically to maintain the proper vehicle safety during navigation.
The JPS, defined as an enhanced A-star algorithm, continuously generates a set of intermediary waypoints $\mathcal{W}_{B,i}^{W}$ towards the user-assigned goal, providing an optimal path that minimizes a desired cost function, as presented in Algorithm~\ref{alg:path_planning}.
To ensure a smooth motion along the path, the waypoints are interpolated with our polynomial-based trajectory approach, which minimizes the snap of the motion, as described in~\cite{jeff2023planner}. Once the drone reaches the final goal position, it hovers and waits for the next command from the user. In the meantime, the user reviews the updated area mesh representation and, if needed, assigns a new path $\bm{\sigma}_{W_v}$ based on the latest collected data. 

\begin{figure}[!t]
  \centering
  \includegraphics[width=0.95\linewidth]{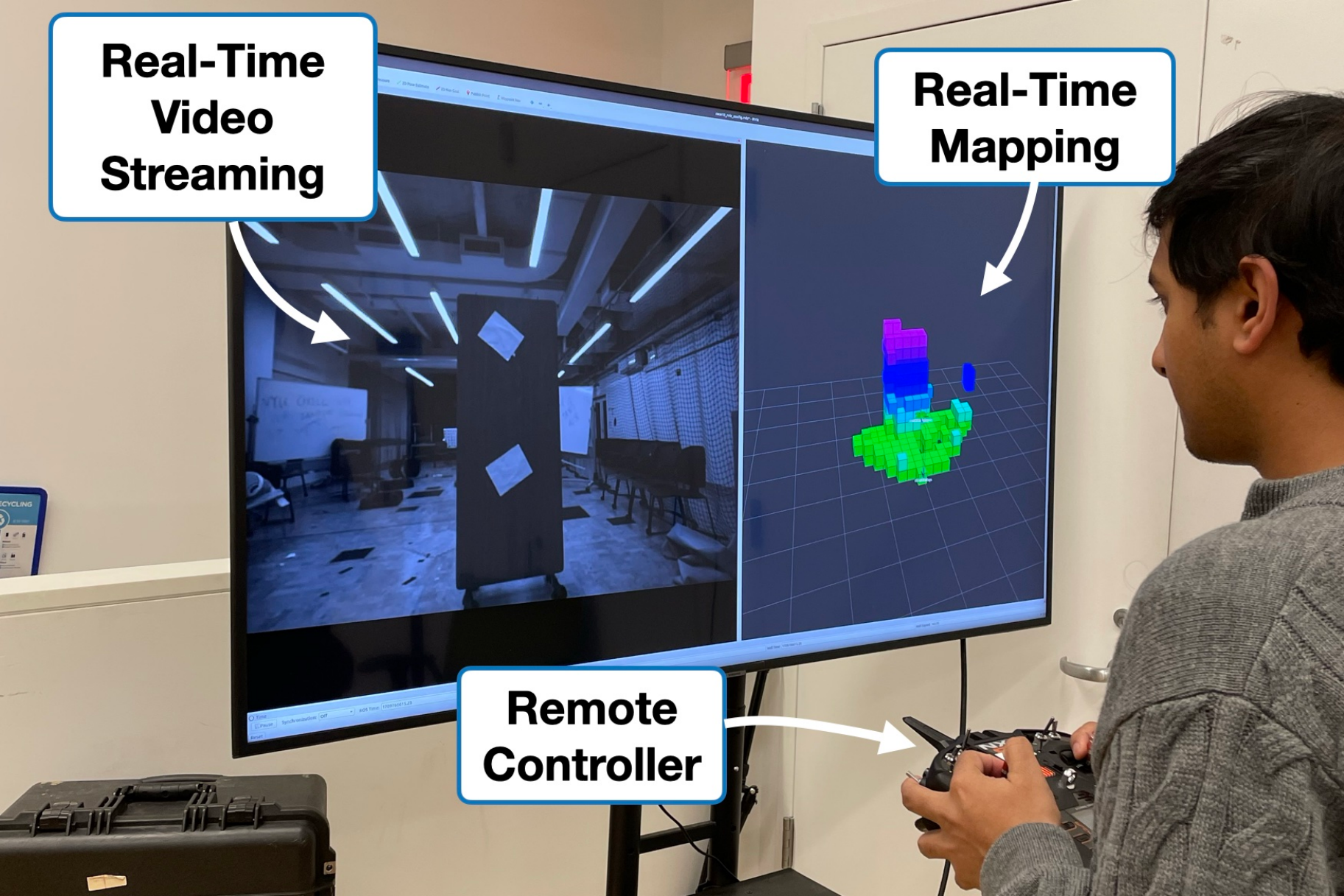}
  \caption{One of the users engaged in the First-Person View modality for robot control.\label{fig:HololensvsFPV}}
  \vspace{-15pt}
\end{figure}

\subsubsection{First Person View (FPV) Navigation}
\label{sec:FPV_navigation}
In contrast to the MR-assisted and autonomous navigation method presented in Section \ref{sec:assisted_nav}, we implement a second interaction approach for a comparative analysis on multiple subjects conducted detailed in Section \ref{sec:experimental_results}. 

In this mode, users operate without assistance from MR representations or the drone's autonomous obstacle avoidance capabilities. They have complete control of the drone, which can receive velocity commands from a Remote Controller (RC) in FPV mode.  Rather than spatial mapping and MR visualization for robot status feedback, we opt for a traditional front-view camera streaming of the current robot view to the user and real-time 3D mapping displayed on a flat TV monitor as shown in Fig. \ref{fig:HololensvsFPV}. To improve control intuitiveness, the combined action of the drone control stack and localization algorithm maintains the drone in a stable hovering condition upon receiving any user command.
\section{Experimental Results}~\label{sec:experimental_results}
 \vspace{-15pt}
\subsection{System Setup}
~\label{sec:System_Setup}
% \begin{figure}[!t]
%   \centering
%   \includegraphics[width=0.95\linewidth]{figs/FPV.png}
%   \caption{One of the users engaged in the First-Person View modality for robot control.\label{fig:HololensvsFPV}}
%   \vspace{-15pt}
% \end{figure}
In the experiments, we use a custom-designed aerial robot called RACE with a total weight of $1.308~\si{kg}$, which uses a PixRacer Pro flight controller and an NVIDIA Jetson Xavier NX as the central processing unit for position and high-level control running on Ubuntu 20.04 and ROS\footnote{\url{www.ros.org}}. The drone leverages an Intel Realsense D435i stereo camera for localization using a customized version of OpenVINS~\cite{openvins}, which estimates the robot states at $100~\si{Hz}$.
For autonomous path planning, the JPS algorithm employs an occupancy grid map of $20~\si{m}\times 20~\si{m}\times 20~\si{m}$ with $0.2~\si{m}$ voxels' side length. The occupancy grid map is constructed based on the point cloud $P_{robot}^{W}$ information received from the drone camera.
We select the Microsoft HoloLens 2.0 as an HMD device~\cite{ungureanu2020hololens} where we upload our custom MR application made in Unity 2020.3.23.f1.
For localization and rigidly overlay virtual objects in the real world, the HMD runs a BAD-SLAM~\cite{schops2019bad} visual-inertial SLAM algorithm at $60~\si{Hz}$.
For communication between the drone and HoloLens, we employ a Ubiquiti Nano5 WiFi operating on a $5$ GHz frequency band with an $80$ MHz channel bandwidth, leveraging a TCP Connection based framework, as depicted in Fig.~\ref{fig:block_diagram}. %Communication between the drone and HoloLens 2.0 was established through a ROS-based TCP network, as depicted in Fig.~\ref{fig:block_diagram}. 
%This network helps in the real-time exchange of data between the drone and HoloLens, including pose, point cloud, meshes, and trajectory information, thereby generating accurate relative virtual representations of each device. 

\subsection{User Task Description}
\vspace{-10pt}

\begin{figure}[!t]
  \centering
  \includegraphics[width=1\columnwidth]{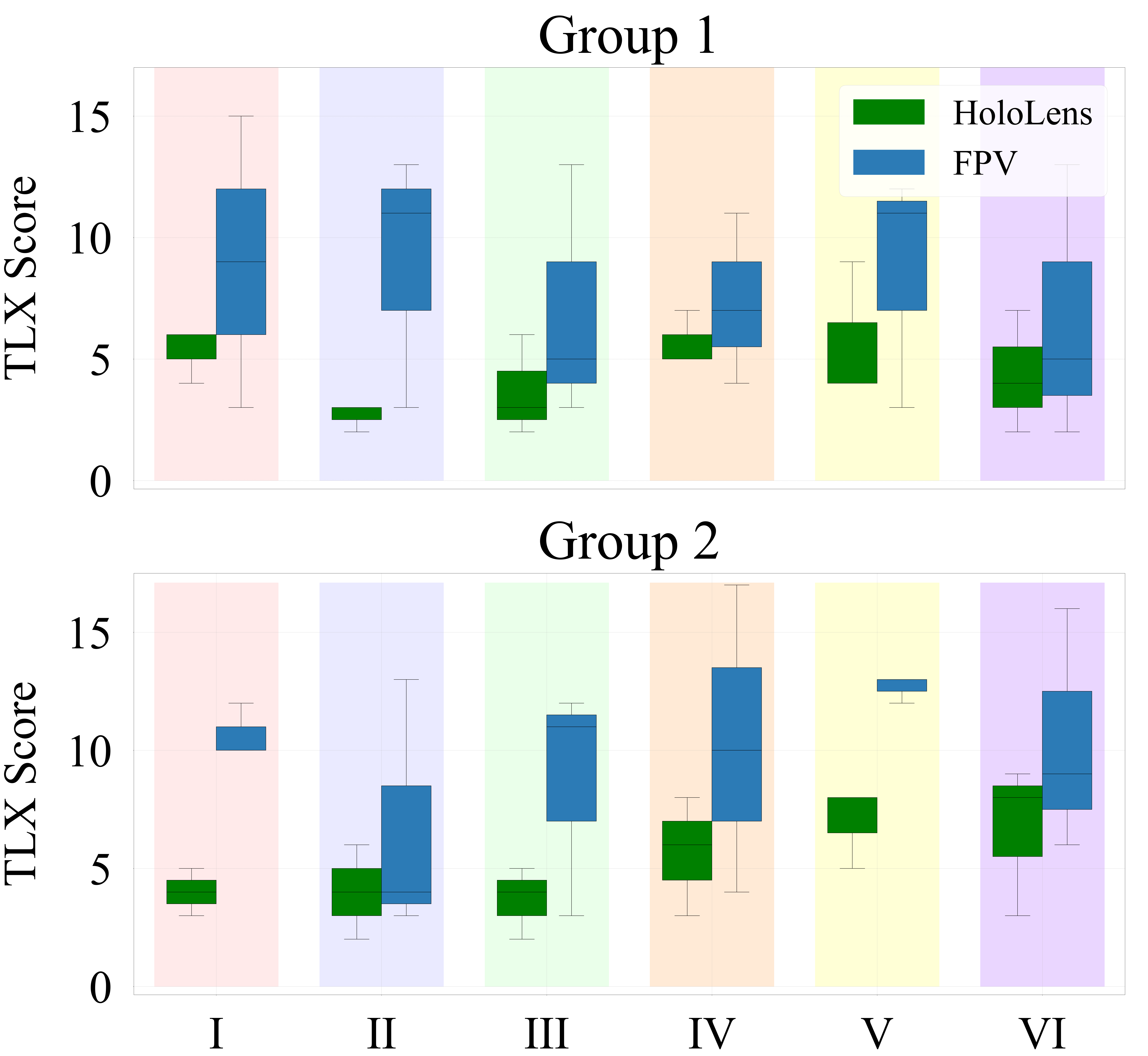}
  \caption{NASA TLX Feedback of participants obtained on subjects belonging in Group $1$ (Expert) and  Group $2$ (Novice). Labels I, II, III, IV, V, and VI denote Mental Demand, Physical Demand, Temporal Demand, Performance, Effort, and Frustration. \label{fig:human_tlx_feedback}}
    %\vspace{-15pt}
\end{figure}

\begin{figure}[!t]
  \centering
  \includegraphics[width=\columnwidth]{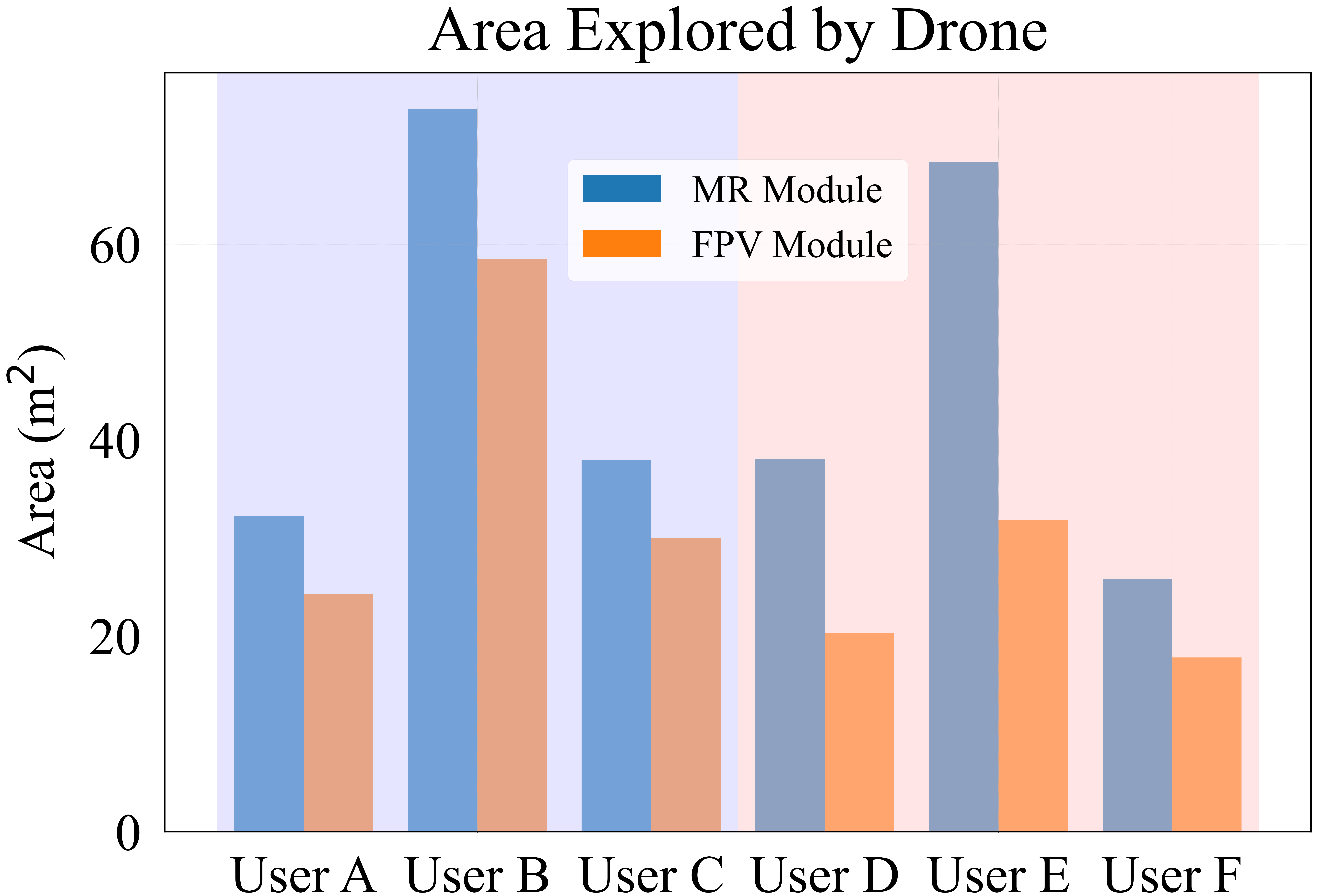}
  \caption{Area navigated by drone during MR and FPV modality. The blue area denotes expert Group $1$ users while the red area denotes novice Group $2$ users.\label{fig:hmi_area_covered}}
    \vspace{-10pt}
\end{figure}
~\label{sec:task_description}

To validate the effectiveness of the proposed solution over the First Person View with onboard localization modality generally preferred in cluttered and GPS denied environments, we test the system assisting the user in an exploration scenario task on $6$ different subjects.  
We divide subjects into two groups: \textit{Group 1}, consisting of individuals with expertise in drone piloting, and \textit{Group 2}, composed of non-expert piloting users. 
% it was mentioned in the review
Each participant performed the exploration task twice, once using the MR modality and once using the FPV modality, with a maximum interaction time of 6 minutes per modality. To account for potential learning effects - where performance could improve in the second trial due to increased familiarity with the task - we counterbalanced the order of the modalities among participants with training sessions and minimize order-related bias in the comparison of MR and FPV effectiveness.
% We ask each subject in both groups to perform the exploration task twice in a cluttered environment, each time using the two modalities described in Section \ref{sec:system_design}. 
% The total interaction time for each modality is $6$ minutes maximum, starting immediately after the robot takes off. 
% Prior to each session, all users are trained in simulation for $20$ minutes to make them familiar with the use of the MR mode and the FPV mode. 
%After the robot autonomously takes-off, it hovers until a new user command is provided in both interaction modalities. 
In Assisted MR navigation, we ask the users to leave the space where the robot is and start exploring the remaining of the building. 
The users decide when and where to provide the desired waypoints $\mathcal{W}_{B,i}^{W}$ or the desired trajectory $\bf{\sigma}_{d_{W}}$ to the robot using the two proposed Tasks Buttons, \textit{Single Waypoint Assignment} or \textit{Multiple Waypoints Assignment}.
In both modes of operation, we have repeatedly informed participants that the primary objective is to maximize the exploration area within the $85~\si{m^2}$, which represents the area of our flying arena.
In the second modality, users cannot simultaneously explore other areas since they can rely only on the real-time camera streaming and mapping displayed on the monitor, as shown in Fig.~\ref{fig:HololensvsFPV}.

\subsection{Results}
\label{results}
\begin{figure*}[!t]
  \centering\includegraphics[width=\textwidth, trim={0 30 0 0},clip]{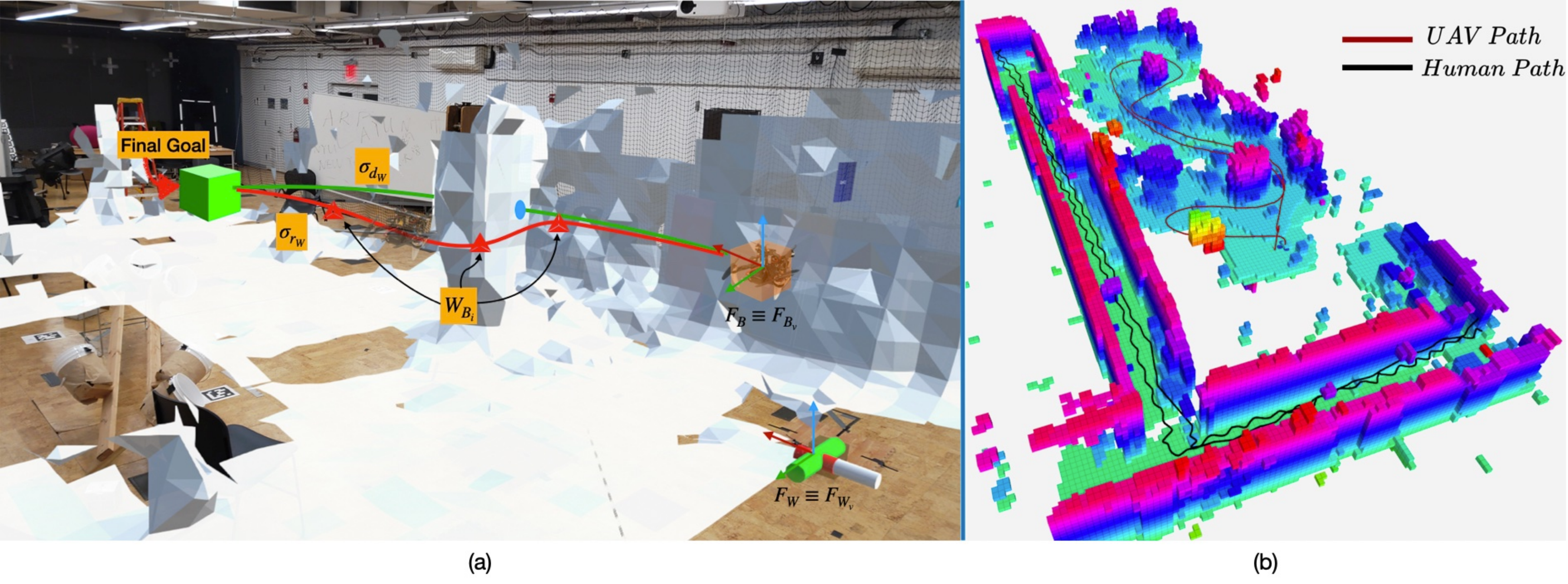}
  \caption{Exploration paths: (a) Real-scale mesh visualization overlaying the environment as visible through the Hololens, with a representation of the re-planned (red) and the user provided trajectory (green) to the robot, b) Bird's-eye view of the resultant octomap and recorded paths of the collaborative human-drone exploration. \label{fig:uav_path_result}}
  \vspace{-10pt}
\end{figure*}
We divide the results into two parts. First, at the end of each interaction type, as a measure of perceived effort, the subjects fill the NASA TLX \cite{hart1988development} module. 
%This data gave insight into the user experience provided by our MR-based system compared to the traditional FPV approach. 
Second, we couple the results obtained from the users' TLX examination scores with a quantitative measurement of the explored area given the size of the proposed environment. %We identify insightful results related to the ability of the operator to safely maneuver the drone within the mission area.
%collecting as much as information as possible from the environment. 
As a metric analysis, we formulate the hypothesis that with the MR reality support, users can explore a larger area in the same time of $6$ minutes, therefore reducing their cognitive effort regardless of their piloting experience, compared to the FPV piloting technique. Qualitative results of the covered mapped area with an example of $\bf{\sigma}_{d_{W}}$ and $\bf{\sigma}_{r_{W}}$ paths, obtained after the re-planning phase, are shown in Fig.~\ref{fig:hmi_area_covered}.  To ensure a smooth interaction, we optimize the average communication latency between the drone and HoloLens to $2.0~\si{ms}$, with peaks of $3.0~\si{sec}$ for more complex spatial data.

\subsubsection{TLX Results} 
\label{TLX}
\begin{table}[!t]
\setlength\tabcolsep{0pt} % make LaTeX figure out intercolumn spacing

\begin{tabular*}{\columnwidth}{@{\extracolsep{\fill}} ll cccc}
\toprule
  Modality & \multicolumn{3}{c}{TLX Scores} \\
\cmidrule{2-4} 
& Group 1 & Group 2 & Overall \\
\midrule
     MR & $24.33 \pm 4.98$ & $29.33 \pm 5.31$ & $26.83 \pm 5.72$\\ 
     FPV & $46.33\pm  19.36$ & $57.33 \pm 10.14$ & $51.83 \pm 16.4$ \\

\bottomrule
\end{tabular*}
\caption{Comparison of the NASA TLX average values between expert Group $1$, novice Group $2$ and Overall.}
\label{table:1}
\end{table}

\begin{table}[!t]

\setlength\tabcolsep{0pt} % make LaTeX figure out intercolumn spacing

\begin{tabular*}{\columnwidth}{@{\extracolsep{\fill}} ll cccc}
\toprule
  Modality & \multicolumn{3}{c}{Average Surface Discovered [\si{m^2}]} \\
\cmidrule{2-4} 
&  Group 1 & Group 2 & Overall \\
\midrule
     MR & $48.02 \pm 14.93$ & $44.07 \pm 17.9$ & $46.04 \pm 16.41$\\ 
     FPV & $37.58 \pm 18.39$ & $23.33 \pm 6.12$ & $30.45 \pm 12.25$ \\

\bottomrule
\end{tabular*}
\caption{Comparison of the average values of mapped surface between expert Group $1$, novice Group $2$ and Overall.}
\label{table:2}
\vspace{-10pt}
\end{table}

The average TLX ratings for both groups are illustrated in Fig.~\ref{fig:human_tlx_feedback} and in Table~\ref{table:1}. The former provides a visualization of the different metrics adopted in the TLX during the two modalities among the different groups, whereas the latter  
shows the average NASA TLX score and the related standard deviation obtained during the same test.  
In particular, it is noticeable how the MR modality, operated through the Hololens, obtains almost a lower global value in all of the categories of the NASA TLX when compared to the proposed standard FPV operation modality across the entirety of the subjects. In particular, we can observe a score reduction of $47.48\%$ in Expert Group $1$ and $48.84\%$ in Novice Group $2$ with an overall reduction of $48.23\%$.
%From Fig.~\ref{fig:human_tlx_feedback} is visible that the TLX score for Group 1 for the MR modality is 24.33 \pm 4.98. The average of the total score for the FPV modality is 46.33\pm  19.36. Thus, the total TLX score reduced by $47.48\%$ when the user where executing the session with the MR modality, as compared to the standard method on average. 
%A similar results is obtained in Group 2. The average of the total TLX score for the MR modality is 29.33 \pm 5.31. For the FPV modality, the average of the total TLX score is 57.33 \pm 10.14. Thus, the interaction with the Hololens, after an appropriate training session, decreased the user mental load by $48.84\%$ compared the proposed standard method on average. 
%Overall, the average of the total TLX score for the MR modality was 26.83, with a standard deviation of 5.72. For the FPV modality, the average of the total TLX score was 51.83 with a standard deviation of 16.4. We recorded that the subjects mental burden during the Mixed Reality session has been decreased by $48.23\%$ compared to the FPV piloting. 
These results confirm the expectations defined in Section~\ref{sec:task_description}, highlighting how the use of unbounded MR applications 
%provide to the user insightful information regarding the robot mapping status, that once linked with the proposed gestured based trajectory drawing and the autonomous navigation capabilities of the robot, 
decrease consistently the user's cognitive effort and improve the quality of interaction and the possibility of success of the mission, independently from the users' background.  

\subsubsection{Mapping Exploration Results} 
\label{sec:mapping}
In parallel to the TLX results, we collect spatial information to assess the quality of the exploration depending on the type of interaction and the background of the users in the two groups. 
The results from interactions using the proposed modalities are presented in Fig.~\ref{fig:hmi_area_covered} and Table~\ref{table:2}. The figure offers a direct comparison of the explored drone area for each participant using the two approaches, while the table displays the average explored area for expert Group 1, novice Group 2, and overall. Despite participants' flight proficiency and prior drone piloting experience, Fig.~\ref{fig:hmi_area_covered} clearly demonstrates that MR-assisted navigation coupled with drone autonomous flight capabilities outperforms classic FPV control in terms of explored area. In expert Group $1$, consisting of skilled pilots, the MR-assisted approach enhances the explored area by approximately $27.75\%$, showcasing improved performance compared to the standard FPV technique. Similarly, in novice Group $2$, less skilled users exhibit significant improvement using MR visualization, experiencing an $88.9\%$ increment in the explored environment area, even after the training session with the FPV interaction. This enhancement is coupled with a noticeable reduction in users' effort workload, as detailed in Table~\ref{table:1} TLX scores. Overall, the proposed MR-based spatial assisted navigation yields a $51.15\% $ increase in the explored area from the robot only.
Moreover, during the MR modality, participants further simultaneously explore an average area of $108~\si{m^2}$, leveraging the mapping capabilities enabled in the HoloLens. This augmented exploration is not possible in the FPV modality, where the participant is confined to the drone's perspective and its constant monitoring. 

Finally, in Fig.~\ref{fig:uav_path_result} we provide a combined representation of the a) full-scale virtual map representation visualized by the user through the MR headset, overlaying the real world, and b) the visualization of the user and robot combined octomaps, obtained by the users during the experimental session.
Moreover in Fig.~\ref{fig:uav_path_result}.a additional holographic details are presented like the desired user forwarded trajectory $\bf{\sigma}_{d_{W}}$, passing through an obstacle, and the drone re-planned one $\bf{\sigma}_{r_{W}}$, passing through multiple waypoints $\mathcal{W}_{B,i}^{W}$ leading the robot safely to the final goal, which is represented by the green cube.
As additional information to the reader, Fig.~\ref{fig:uav_path_result}.b also visualizes the human walked trajectory, recorded using the Hololens 2 embedded sensors.

%we analyzed the impact of network latency on performance. The communication framework maintains an average latency of 2.0 ms, with a peak of $3.0~\si{sec}$ for more complex spatial data. These delays affected MR visualization updates but had minimal impact on autonomous flight stability due to local redundancy mechanisms.

%Finally, a representation of the final user environment map using the MR interface proposed in Section~\ref{sec:system_design} is visible in Fig.~\ref{fig:uav_path_result}. 
%In particular, Fig.~\ref{fig:uav_path_result}.a provides a combined representation of the area mapped by the user (red point cloud) linked with a geometric mesh representation of the area mapped by the robotic agent, depicted on the user side as visualized in Fig.~\ref{fig:intro}.

%Moreover, in Fig.~\ref{fig:uav_path_result}.b and  Fig.~\ref{fig:uav_path_result}.c respectively the human and the robot trajectories followed during the area inspection are visualized. As observable, the robot trajectory $\bm{\sigma}_{W_v}$ is cleared from obstacles. This results are achieved only using the re-planning technique discussed in Section \ref{sec:system_design}. This delivers to the autonomy pipeline the duty to leverage the robot perception and to re-plan a collision free safe path, with at least  $0.4~\si{m}$ clearance from the obstacles. 

%!TEX root = ARTICLE.tex
\section{Conclusion}~\label{sec:conclusion}
In this paper, we introduce a novel MR human-drone interactive paradigm for drone assisted and collaborative navigation in indoor unstructured environments. By leveraging a shared map between the user and the robot, the user gains enhanced spatial awareness of the environment, enabling the planning of safe, obstacle-free paths for the drone even when it is outside the user’s line of sight. In this way, despite the two agents are not being spatially co-located, the humans can rely on the drone's autonomous capabilities for solving complex and dangerous tasks such as exploration, search and rescue, and inspections. Our results, supported by a user case study, validate the technical effectiveness of the approach, demonstrating a reduction in cognitive effort regardless of the user's piloting experience. Future works will aim to provide a cloud MR interface democratizing the setup access. Finally we aim to extend the user case study to more subjects.  

%In this paper, we presented a novel human-drone interactive interface paradigm based on MR for drone assisted and collaborative navigation in an unstructured environments. By leveraging a combined user and robot map, the user can increase the spatial awareness of the environment to establish safe and obstacle free path to the drone, even if this is not within the users' line of sight. In this way, despite the two agents not being spatially co-located, the humans can rely on the drone's autonomous capabilities for solving complex and dangerous tasks such as exploration, search and rescue, and inspections. Our results, including $6$ users tests, prove the validity of the technical elements of the approach, even when compared with another piloting technique.
%In the future, we plan to perform additional user case studies, trying to extract  statistical relevant patters on the proposed metrics to analyze the users' behaviors. Additionally, we aim to provide a cloud MR interface democratizing the setup access. Finally, we aim to develop a comprehensive multi-user/multi-robot interactive system.

% \todo{adjust this param here}
\addtolength{\textheight}{0.0cm}   % This command serves to balance the column lengths

\bibliographystyle{IEEEtran}
\bibliography{references}

\end{document}